\documentclass[letterpaper, 10 pt, conference]{ieeeconf}  

\IEEEoverridecommandlockouts                              

\usepackage{multirow}
\usepackage{float}
\usepackage{tabularx}
\usepackage{array}
\usepackage{makecell}
\usepackage{amsmath}
\usepackage{graphicx}	
\usepackage{tabu}                      
\usepackage{booktabs}                  
\usepackage{hyphenat}
\usepackage{url,hyperref,lineno,microtype,subcaption}
\usepackage{tikz}
\def\checkmark{\tikz\fill[scale=0.4](0,.35) -- (.25,0) -- (1,.7) -- (.25,.15) -- cycle;} 

\title{\LARGE \bf
Does elderly enjoy playing Bingo with a robot? A case study with the humanoid robot Nadine}

\author{Nidhi Mishra$^{1}$, Gauri Tulsulkar$^{2}$, Hanhui Li$^{3}$ and Nadia Magnenat Thalmann$^{4}$ 
\thanks{$^{1}$ Nidhi Mishra is with Institute for Media Innovation, Nanyang Technological University, 50 Nanyang Drive, Singapore 
        {\tt\small nidhi.mishra@ntu.edu.sg}}
\thanks{$^{2}$ Gauri Tulsulkar is with Institute for Media Innovation, Nanyang Technological University, 50 Nanyang Drive, Singapore 
        {\tt\small grtulsulkar@ntu.edu.sg}}
\thanks{$^{3}$ Hanhui Li is with Institute for Media Innovation, Nanyang Technological University, 50 Nanyang Drive, Singapore
        {\tt\small hanhui.li@ntu.edu.sg}}
\thanks{$^{4}$ Nadia Magnenat-Thalmann is with Institute for Media Innovation, Nanyang Technological University, 50 Nanyang Drive, Singapore and MIRALab, University of Geneva, Switzerland
        {\tt\small nadiathalmann@ntu.edu.sg}}
}

\begin{document}
\maketitle
\thispagestyle{empty}
\pagestyle{empty}

\begin{abstract}

There are considerable advancements in medical health care in recent years, resulting in rising older population. As the workforce for such a population is not keeping pace, there is an urgent need to address this problem. Having robots to stimulating recreational activities for older adults can reduce the workload for caretakers and give them time to address the emotional needs of the elderly. In this paper, we investigate the effects of the humanoid social robot Nadine as an activity host for the elderly. This study aims to analyse if the elderly feels comfortable and enjoy playing game/activity with the humanoid robot Nadine. We propose to evaluate this by placing Nadine humanoid social robot in a nursing home as a caretaker where she hosts bingo game. We record sessions with and without Nadine to understand the difference and acceptance of these two scenarios. We use computer vision methods to analyse the activities of the elderly to detect emotions and their involvement in the game. We envision that such humanoid robots will make recreational activities more readily available for the elderly. Our results present positive enforcement during recreational activity, Bingo, in the presence of Nadine.

\end{abstract}


\section{Introduction}
Many nursing and elderly homes face the challenges of balancing costs and quality due to an increased demand for long-term care for the elderly. Besides, most nursing homes focus on medical and nursing procedures because they lack human resources and expertise to address the psychosocial well-being demands. As such, there is greater demand for quality services and resources, in addition to the existing challenges of human-resource shortages, limited expertise, and rising costs of healthcare and social care. To bridge the gap caused by resource shortages, we propose to deploy a social robot in nursing homes, taking advantage of AI technologies. 

This research aims to apply the human-robot interaction (HRI) technology to draw the attention of the elderly and stimulate their interest.  As discussed in ~\cite{correia2017social} entertainment by robots constitutes a relevant and promising area of application in HRI that needs to address many different populations, including older adults. ~\cite{agrigoroaie2018physiological}, and ~\cite{agrigoroaie2018outcome}  discuss those game activities with robots that can help the cognitive stimulation of the elderly. Bingo is one such fun and popular activity that triggers long-term memory, making it one of the more stimulating brain games for the elderly. In this paper, we like to present a human-robot interaction study where a Humanoid Social Robot Nadine ~\cite{NadineSR_CGI_2019} facilitates multiple sessions of Bingo games for a group of elderly. 

Before Nadine was deployed at the nursing home, the nursing home's care staff used to host the Bingo sessions. We recorded two of these sessions as our baseline for the study. We studied several Bingo sessions between professional care staff and the elderly, using them to define appropriate and proactive behaviour in Nadine.  We developed and deployed a module that helped Nadine host the Bingo game. Nadine can carry the game by calling out the Bingo numbers, verifying winning Bingo players, and celebrating with the winners.

This study aims to discover the acceptance and the efficacy of Nadine's features and behaviours by elderly residents in a nursing home during multiple Bingo games. To obtain a comprehensive understanding of the effects of Nadine, we use objective tools for data analysis. The objective tools are based on cutting-edge computer vision techniques, such as Deep Neural Networks (DNNs), to automatically evaluate the emotional states and the residents' quality of engagement.

The rest of the paper is organized as follows: We provide related work for humanoid robots assisting in the nursing home. in section \ref{SOA}.  In section \ref{ES}, we explain the experimental setup and adaptation technique of the Nadine social robot at the nursing home. We describe the details of our data collection methods, and we provide details of our framework to analyze the data collected in section \ref{DCA}. In section \ref{res}, we present and discuss the data analysis's experimental results. We provide conclusions in section \ref{CAD}.

\section{Literature Survey}
\label{SOA}
Efforts have been made to enhance the overall mental health of the elderly by providing robots as companions. These robots can help the elderly live in their own homes and communities, safely and independently, by providing assistance or services~\cite{law2019developing}. ~\cite{cohen2009engagement} showed that participants in the nursing homes engaged more frequently in the presence of different types of stimuli, such as moderate levels of sound and small groups of people.

As described in ~\cite{martinez2020socially}, social assistive-robot-based systems with the abilities to perform activities, play cognitive games, and socialize, can be used to stimulate the physical, cognitive, and social conditions of older adults and restrain deterioration of their cognitive state. 

In  ~\cite{thompson2017robot}, ~\cite{li2016user}, ~\cite{louie2020social}, and ~\cite{louie2015tangy}, a robot, Tangy was used to facilitate a multi-player Bingo game with seven elderly residents in a LTC facility. The robot would autonomously call out the bingo numbers and check individual cards of the players to provide help or let them know if they had won Bingo. The results of the studies showed that they had high levels of compliance and engagement during the games facilitated by the robot.  Tangy was humanlike, however the appearance was not that of a  realistic humanoid. The study conducted was for a short duration, which resulted in us being unable to see if the robot could hold the interest over a course of time. Also, the number of participants in each session was low, which raises the question of "if Tangy can hold the interest of a larger group?". 

Other studies such as ~\cite{stevie2020bingo} with Stevie the robot show that these interactions can improve the enjoyability of the games and can be used for cognitive stimulation by playing Bingo for the elderly. The robot, Matilda, was used~\cite{khosla2013embodying} to improve the emotional wellbeing of the elderly in three care facilities with 70 participants in Australia. They found that a Bingo game activity with Matilda positively engaged the elderly and increased their social interaction. Some participants wanted Matilda to participate in all the group activities as she rewarded the winner by singing and dancing. 

Paro robot was placed at nursery homes ~\cite{chang2013use}, and in everyday elderly care ~\cite{vsabanovic2016socializing}. These studies found that the extended use of Paro brought about a steady increase in the physical interaction between the elderly and the robot and also led to a growing willingness among the participants to interact with it. ~\cite{vsabanovic2013paro} also reported an increase in sociability within the group of elderly as well as other social benefits. When Paro took part in group activities, it positively influenced mood change, reduced loneliness, and resulted in a statistically significant increase in interactions.

~\cite{silbot2019blond} presented a user study where Silbot robot was hosting Bingo as part of a brain fitness instructor in an elderly care facility in Denmark. The author had highlighted hardware and software problems and mentioned battery problems,  usability problems, and multiple use case issues. It was studied for two years long, and many of the issues with having for the long term were highlighted, which can be the base of our study.

In the table \ref{Tab: Summary SOA}, we classify the available state of the art robots based on similar studies with robots conducting a recreational activity where '\checkmark' indicates the presence and 'X' indicates the absence of those characteristics.

\begin{table}[h]
\centering    
    \begin{tabular}{cccccccc}
        \toprule
        \textbf{Robot and Paper}  & \textbf{(a)} & \textbf{(b)} & \textbf{(c)} & \textbf{(d)} & \textbf{(e)} & \textbf{(f)} & \textbf{(g)} \\
        \toprule
        Pepper ~\cite{li2016user} & X & X & \checkmark & \checkmark & \checkmark & \checkmark & X \\
        \hline
        Zora ~\cite{melkas2020impacts} & X & X & \checkmark & N/A &  \checkmark  & N/A & X \\ 
        \hline
        Tangy ~\cite{tangy} & X & X & \checkmark & \checkmark & \checkmark & N/A & X \\
         \hline
        Silbot ~\cite{silbot2019blond} & X & \checkmark & \checkmark & \checkmark & \checkmark & N/A & X\\
        \hline
        Zora ~\cite{huisman2019two} & X & X & \checkmark & \checkmark & \checkmark & X & X \\
        \hline
        Zora ~\cite{tuisku2019robots} & X & X & \checkmark & \checkmark & \checkmark & X & X \\
        \hline
        Stevie ~\cite{stevie2020bingo} & X & \checkmark & \checkmark & X & \checkmark & X & X \\
        \hline
        Matilda ~\cite{khosla2013embodying} & X & \checkmark & \checkmark & X & \checkmark & N/A & X \\
        \hline
        Nadine & \checkmark & \checkmark & \checkmark & \checkmark & \checkmark & \checkmark & \checkmark \\
        \bottomrule
    \end{tabular}
    \caption{Summary of the robots for recreational activity for elderly care where (a) Humanoid realistic appearance, (b) Facial expressions, (c) Gestures, (d) IoT, (e) Gazing, (f) Multilingual and (g) Computer Vision based analysis}
    \label{Tab: Summary SOA}
\end{table}

In most studies, robots used are not realistic as Nadine humanoid robot. Nadine can emote natural human communication with its humanlike features, as reported in ~\cite{baka2019talking}. Nadine can give facial expression, respond with gestures, make eye contact with the elderly, and compare these with other robots. Studies have shown that the interaction is stimulating with face and arm movements and could arouse curiosity and interest ~\cite{law2019developing}. We also classify Whether the robot can speak and understand speech; if yes, is it multilingual? Next, for functionalities, we address different aspects of the vision-based capabilities of Robots. We study whether the robot can understand the environment to gaze at the elderly and understand their facial emotions. We examine if previously studied robots could control and interact with the devices (such as TV, speakers, temperature control, etc.) to facilitate the recreational activity. Further, our classification builds upon the analysis method used for the statistical results in previous studies. Specifically, if they were, AI or ML enabled. 

The above are deduced from the table that has not been taken into account in the literature so far. We believe that our study could be a step forward in introducing the humanoid social Nadine robot with humanlike characteristics in appearance and its mimic of human behaviour.

\section{Experimental Setup}
\label{ES}

Nadine is a socially intelligent, realistic humanoid robot with natural skin, hair, and appearance. She has 27 DOF, which enables her to make facial movements and gesticulate effectively, as documented in ~\cite{xiao2014human} and ~\cite{beck2016motion}. Nadine was seated in a ward's activity area in the nursing home, where she also hosts Bingo games and interacts with the elderly.

\begin{figure*}[]
	\begin{center}
			\includegraphics[width=\textwidth]{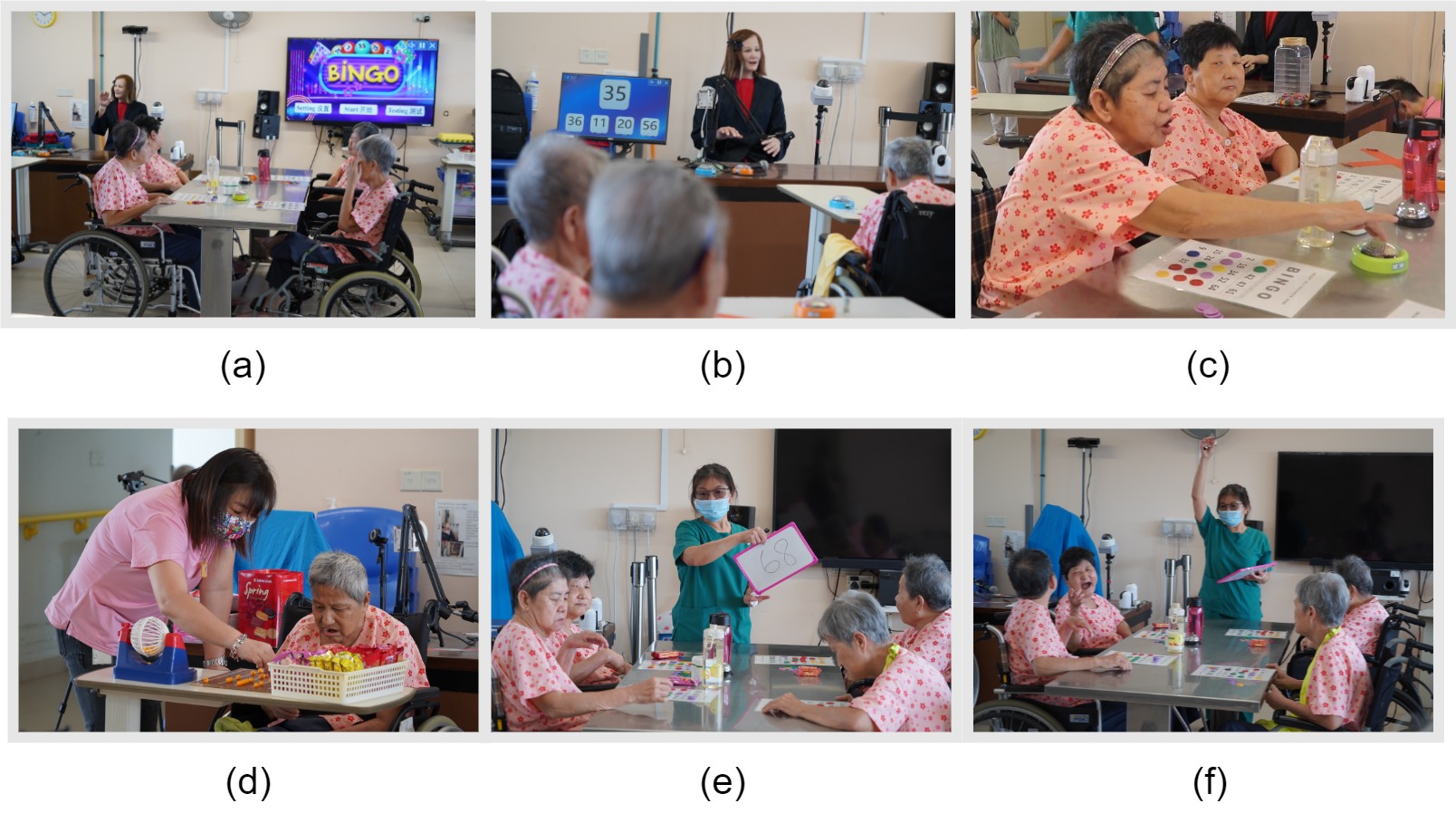}
	\end{center}
	\caption{Nadine hosting bingo session VS Care-staff hosting bingo session. (a) Nadine starting Bingo with Encouragement, (b) Nadine displaying number on the TV, (c) Elderly pressing buzzer on getting Bingo, (d) Care-staff and elderly Start running Bingo, (e) Care-staff going to each elderly table to announce number and (f) Care-staff raising hand for elderly on getting Bingo}
	\label{fig:TherapyStaff}
\end{figure*}

\subsection{Architecture}

 Nadine’s architecture is described in ~\cite{NadineSR_CGI_2019}; it consists of three layers: perception, processing, and interaction. Nadine receives audio and visual stimuli from microphones, 3D cameras, and web cameras to perceive user characteristics and her environment, sent to the processing layer. The processing layer is the core module of Nadine that receives all results about the environment from the perception layer and acts upon them. This layer includes various submodules, such as dialogue processing (chatbot), affective system (emotions, personality, mood), and Nadine's memory of previous encounters with users. Finally, the action/interaction layer consists of a dedicated robot controller, including emotion expression, lip synchronization, and gaze generation.


Nadine can recognize people she has met before and engage in a flowing conversation. Nadine can be considered a part of human-assistive technology ~\cite{magnenat2014social}, as she can assist people over a continuous period without any breaks. She has previously worked at different places that required her to work for long hours ~\cite{mishra2019can}.

\subsection{Adaptation}

The studies ~\cite{neven2010but} and ~\cite{lee2016robot} stated that most robots designed for the elderly do not fulfil the needs and requirements to perform their best. For Nadine to perform her best at nursing homes, we updated some previous models and developed new modules recommended in previous studies~\cite{chang2014exploring}.

\subsubsection{Bingo game}
Before Nadine deployment at the nursing home, care staff use to host bingo sessions for the elderly. They would announce the number and use a small whiteboard to display the number for residents with hearing impairment. To do the same task of hosting the Bingo game, a new module was developed. This module enables Nadine to do the following:
\begin{itemize}
    \item Start the session with greetings and weather information.
    \item Call out the bingo numbers in English and Mandarin.
    \item Call numbers in specific time duration and repetitions.
    \item Enables her to display the current number and four previous number on the TV screen.
    \item The numbers called out are also accompanied by hand gestures, facial expressions, gazing, and background music.
    \item It also allows care staff to control and customize Nadine's Bingo sessions using an attached touch screen. 
    \item Let's Nadine verify winning players and applaud them. 
\end{itemize}
Fabricated in-house buzzers were provided to the residents to press when they win the game. Nadine also played a cheering sound upon confirming a Bingo call from the residents.

\subsubsection{Update in Nadine's existing module}
As a social humanoid robot, Nadine has an emotion engine that controls her emotions, personality, and mood during the interaction, enabling her to perceive the situation (user and environment) and adjust her emotions and behaviour accordingly. As a result, Nadine can generate different emotions such as pleasure, arousal, and dominance. For Nadine to perform best at nursing homes, she needs to appear patient and show no negativity or anger. Therefore, Nadine should exhibit a positive temperament only. A configuration file was set to different parameters that allow Nadine to stay positive and behave accordingly, even when the resident is frustrated, angry or upset with her.

Another important aspect is to reveal positive emotions in Nadine's speech synthesis output. This mainly relates to changing the pitch, tone, and speed modulations. We modified the speech synthesizer to adapt speech output so that Nadine speaks slower and louder and in a low tone to make it easier for residents to understand her.

\subsection{Participants}
Twenty-nine participants aged 60 years above participated in our research. The experiment took place at Bright Hill Evergreen Nursing Home in Singapore. NTU Institutional Review Board has approved this study. A detailed consent form was signed before the onset of the procedure, followed by a detailed explanation of the experiment. We ensured that our participants had no previous experience with robots or any advanced technology. All the sessions were monitored by nursing home care staff. Overall, we recorded 24 sessions with Nadine hosting Bingo and 2 with care-staff hosting Bingo.

\section{Data Collection and Analysis}
\label{DCA}

\begin{figure*}[]
	\begin{center}
			\includegraphics[width=\textwidth]{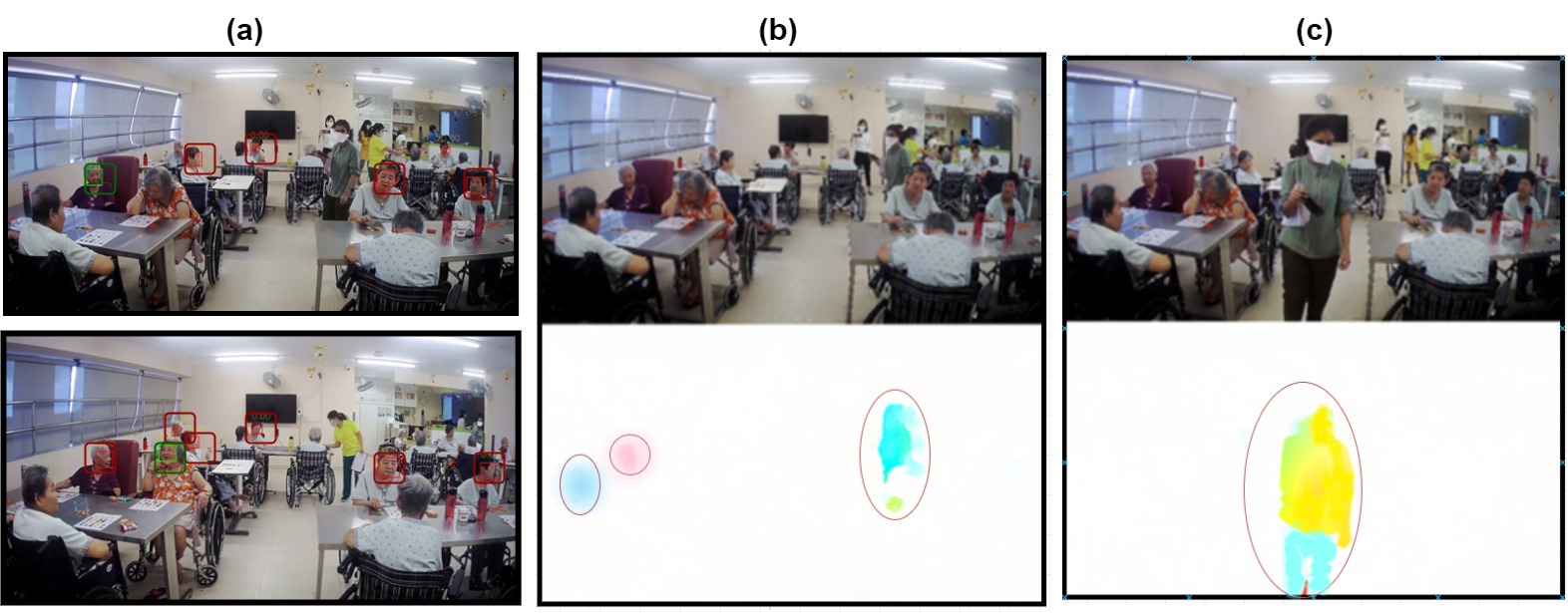}
	\end{center}
	\caption{Example results from computer vision methods. (a) Elderly faces detected in red box with smiling faces in green box, (b) Elderly body movement in three small patches and (c) staff movement in one patch}
	\label{fig:CVresults}
\end{figure*}

To fully comprehend the effects of Nadine's presence in the nursing home, the whole session was recorded by five cameras from different angles. Objective tools based on cutting-edge computer vision techniques were considered. We used deep neural networks (DNNs) to evaluate the emotional and physical states of the elderly automatically. The following three evaluation metrics were focused on: 
\begin{itemize}
    \item \emph{Happiness}, which is the satisfaction level of residents during the Bingo game.
    \item \emph{Movement}, which reflects Nadine's effect on the physical movement of the elderly during the game.
    \item \emph{Activity}, which is the overall care staff physical movement during the Bingo game hosted by Nadine.
\end{itemize}

To provide a quantitative analysis concerning these three metrics, the advantages of DNNs in efficient video processing were exploited. Particularly, we applied four different networks for this study: a face detector\footnote{\url{http://dlib.net/}}, an expression recognizer, an action detector\footnote{\url{https://github.com/open-mmlab/mmaction2}} and an optical flow estimator~\cite{raft2020}. The roles and detailed implementations of these networks are as follows:

The face detector estimates the locations of faces in a given frame. Moreover, as all residents are in wheelchairs, their relative locations can be inferred based on their face locations. Also, only elderly faces are detected since the nursing care staff were wearing medical masks throughout their intervention in Nadine's Bingo sessions. We adopted the Dlib library with its pre-trained convolutional neural network (CNN) to implement the face detector.

The expression recognizer categorizes the expression of a detected face. We consider two classes in our case, i.e., smiling and neutral, and constructed a CNN with ResNet-50~\cite{ResNet} as the backbone. The expression recognizer is trained on the CelebA dataset ~\cite{celebA2018} until convergence.

The action detector is a method to measure the motions and actions of the detected faces. It generates action proposals, which are the locations and confidences of detecting an action. We implement the action detector based on the pre-trained temporal segment network~\cite{TSN} provided in the MMAction2 library. The action detector informs us of the movement of the elderly during the Bingo game session as the face detector detected only their faces.

The optical flow estimator aims at discovering moving targets. We propose to estimate dense optical flow via the recurrent all-pairs field transformation network. At this moment, for an arbitrary region in each frame, the average magnitude of the estimated optical flow in the region can be used to measure the intensity of movement of care staff during the Bingo game sessions.

With the above DNNs, we are now ready to define the quantitative measures of \emph{Happiness}, \emph{Movement} and \emph{Activity}:

\emph{Happiness} is closely related to expressions of smiling and laughing.  Hence, given a target video of  $L$ frames we define group happiness $h$ as:
\begin{equation}
h = \frac{1}{L}\sum\limits_{l = 1}^L {\frac{1}{{{n_l}}}} \sum\limits_{t = 1}^{{n_l}} {{p_t}},
\end{equation}
where $p_t \in [0, 1]$ denotes the probability of the $t$-th detected face in $l$-th frame belonging to the smiling class, which is estimated by the expression recognizer.

\emph{Movement} of elderly is defined using action detector in the following equation where $d_t$ is the confidence of detecting an action:
\begin{equation}
b = \frac{1}{{L}}\sum\limits_{l = 1}^L {\frac{1}{{{n_l}}}\sum\limits_{t = 1}^{{n_l}} {{d_t}} } 
\end{equation}
 
For \emph{Activity} of the care staff, we use the optical flow estimator as following equation where $o_t$ is the average magnitude of optical flow:
\begin{equation}
f = \frac{1}{{L}}\sum\limits_{l = 1}^L {\frac{1}{{{n_l}}}\sum\limits_{t = 1}^{{n_l}} {{o_t}} } 
\end{equation}

Using the above analyses(example shown in figure \ref{fig:CVresults}), we obtained the data for every video across all sessions and further studied it using statistical methods to get meaningful comparisons.  

\section{Results}
\label{res}

In order to determine whether the presence of Nadine hosting bingo games has any effect on the elderly, the video material is analyzed. Four aspects of the video material (smile, neutral, body score, optical flow) were compared between the sessions in which Nadine hosted the games and the sessions in which the caretakers hosted the games. After cleaning the footage, there were 24 sessions with 29 elderly participating in each game in which Nadine was the host and 2 sessions in which the caretakers hosted the games. For each of the sessions, there were multiple camera angles that generated the footage. The analyses were conducted by comparing all the footage from each of the available cameras without compressing them into single averages for the sessions.

To compare the scores for when Nadine hosted the games and the sessions in which the caretakers hosted the games, an independent samples t-test was conducted. The test results can be seen in table \ref{tab:3}, while the means and the standard deviations of the four variables in the two situations can be seen in table \ref{tab:2}. Before conducting the t-tests, Levene’s test of equality of variance was conducted to check for the assumption of homoscedasticity. It was determined that two of the variables had unequal variance (variables smile and body score). Therefore, for these two variables, modified degrees of freedom were used.

\begin{table}[h]
\resizebox{\columnwidth}{!}{
\begin{tabular}{|c|c|c|c|c|}
\hline
\textbf{Variable}             & \textbf{Situation} & \textbf{Mean   } & \textbf{\thead{Standard \\ Deviation}} & \textbf{\thead{SD Error \\ Mean}} \\ \hline
\multirow{2}{*}{Smile}        & Nadine present     & 0.019         & 0.015                   & 0.002                    \\ \cline{2-5} 
                              & Nadine absent      & 0.010         & 0.006                   & 0.002                    \\ \hline
\multirow{2}{*}{Neutral}      & Nadine present     & 0.947         & 0.179                   & 0.024                    \\ \cline{2-5} 
                              & Nadine absent      & 0.990         & 0.006                   & 0.002                    \\ \hline
\multirow{2}{*}{Body Score}   & Nadine present     & 0.154         & 0.053                   & 0.007                    \\ \cline{2-5} 
                              & Nadine absent      & 0.172         & 0.015                   & 0.005                    \\ \hline
\multirow{2}{*}{Optical Flow} & Nadine present     & 0.181         & 0.133                   & 0.018                    \\ \cline{2-5} 
                              & Nadine absent      & 0.288         & 0.081                   & 0.029                    \\ \hline
\end{tabular}}
\caption{Descriptive statistics of the relevant variables in the two situations.}
\label{tab:2}
\end{table}

\begin{table}[h]
\resizebox{\columnwidth}{!}{
\begin{tabular}{|c|c|c|c|c|}
\hline
\multirow{2}{*}{\textbf{Variable}} & \multirow{2}{*}{\textbf{t}} & \multirow{2}{*}{\textbf{df}} & \multirow{2}{*}{\textbf{Sig. (2-tailed)}} & \multirow{2}{*}{\textbf{Mean Difference}} \\
 &  &  &  &  \\ \hline
\textbf{Smile} & 3.341 & 23.122 & 0.003 & 0.009 \\ \hline
\textbf{Neutral} & -0.678 & 63 & 0.500 & -0.043 \\ \hline
\textbf{Body score} & -2.123 & 38.275 & 0.040 & -0.019 \\ \hline
\textbf{Optical flow} & -2.211 & 63 & 0.031 & -0.107 \\ \hline
\multirow{2}{*}{\textbf{Variable}} & \multicolumn{2}{c|}{\multirow{2}{*}{\textbf{Std. Error Difference}}} & \multicolumn{2}{c|}{\textbf{95\% Confidence Interval of the Difference}} \\ \cline{4-5} 
 & \multicolumn{2}{c|}{} & \textbf{Lower} & \textbf{Upper} \\ \hline
\textbf{Smile} & \multicolumn{2}{c|}{0.003} & 0.004 & 0.015 \\ \hline
\textbf{Neutral} & \multicolumn{2}{c|}{0.064} & -0.171 & 0.084 \\ \hline
\textbf{Body score} & \multicolumn{2}{c|}{0.009} & -0.036 & -0.001 \\ \hline
\textbf{Optical flow} & \multicolumn{2}{c|}{0.049} & -0.204 & -0.010 \\ \hline
\end{tabular}
}
\caption{Results of the t-tests.}
\label{tab:3}
\end{table}

\begin{figure}[h!]
	\begin{center}
			\includegraphics[width=\columnwidth]{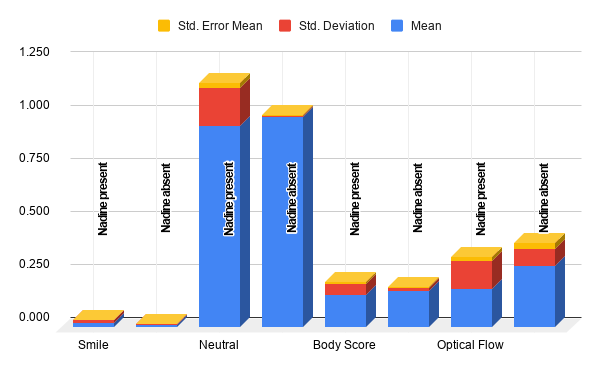}
	\end{center}
	\caption{Means and 95\% CI of measurements.}
	\label{fig:Means}
\end{figure}

As can be seen from table \ref{tab:3}, three of the variables showed significant differences between the two situations: the smile variable was significantly higher in the Nadine group, while the body score and the optical flow variables were higher in the no-Nadine group. There were no differences in the neutral measurement between the two groups. This can also be seen from the figure \ref{fig:Means}, which shows the means and the 95\% confidence errors (error bars).

In order to determine whether or not the reactions of the elderly changed through time, bivariate correlations were calculated between the four variables and the serial number of the session. The higher the serial number is, the later the session was, so a correlation would imply a linear change in the variables with time. The results of the correlational analysis are presented in table \ref{tab:4}.

\begin{table}[h]
\resizebox{\columnwidth}{!}{
\begin{tabular}{|c|c|c|}
\hline
\textbf{Variable} & \textbf{Pearson Correlation} & \textbf{Significance} \\ \hline
\textbf{Smile} & 0.249 & 0.062 \\ \hline
\textbf{Neutral} & -0.241 & 0.071 \\ \hline
\textbf{Body Score} & -0.069 & 0.609 \\ \hline
\textbf{Optical Flow} & 0.113 & 0.403 \\ \hline
\end{tabular}}
\caption{Bivariate correlations between serial number of the session and the four variables.}
\label{tab:4}
\end{table}


\section{Conclusion and Discussion}
\label{CAD}

In this study, the effects of a humanoid robot, Nadine, on the behaviour of elders in a nursing home were investigated. Nadine humanoid social robot was placed at a nursing home where she hosted bingo game for a group of elderly.  The sessions and tracked through video material obtained from several different angles in the nursing home. From the material, four variables were measured: Smile (how many people are smiling per frame and with what intensity), neutral (how many people are sitting neutral per frame and with what intensity), body score (how many people are making body movements per frame and with what intensity), and optical flow (the overall flow in the environment per frame and its intensity). 

The results indicated three significant differences in the bingo sessions in which Nadine was present compared to those in which she was not. The elderly was smiling more, they were moving around less, and the optical flow, which primarily relates to how many nursing home's care staff had to move, was also lower. Therefore, it can be concluded that the situation in the nursing home was better when Nadine was present: the residents were calmer and happier, while the staff had less work to do. This is in line with previous research [7-11], all of which have shown the potential positive effects of employing robotic assistance in nursing homes. Since Nadine has a humanoid appearance and the ability to communicate, read and exert facial expressions related to emotions, it is logical to assume that her presence can improve the states of the residents of nursing homes while simultaneously unloading the nursing home staff. This study confirms that this may be the case since all the changes in the measured variables between the situations in which Nadine was present and those in which she was not point in that direction. 

On the other hand, there were no significant changes in the variables through time. This may be due to the nature of the variables: since the effects of Nadine are that the elderly is more concentrated, happier, and less needy (seen through the lower movement of the staff), it is plausible that these effects did not change over time. Furthermore, this means that the positive effects of Nadine's presence can be seen very early and that no period of adaptation is needed to achieve the changes. Therefore, her presence can bear immediate positive changes in a nursing home or a similar facility.

The realism of Nadine's appearance and interactions are of paramount importance for her usage in human interactions, especially amongst the elderly. Since they are the least adopted and used to technology, it is beneficial to bridge this gap by using robots who look like humans. In these situations, all the benefits of having people do a particular job can be combined with the benefits of using robots, which leads to the best possible outcomes for both users and organizations. That is why research on humanoid robots is so important and why it needs to be progressed further. This study, which showed that the usage of Nadine in a bingo sessions setting could be very beneficial to the residents of the nursing home, is a step in that direction. Future studies should continue to investigate these issues and determine all the settings in which humanoid robots' usage could be beneficial.

\section*{ACKNOWLEDGMENT}
This research is partly supported by the National Research Foundation, Singapore under its International Research Centres in Singapore Funding Initiative, and Institute for Media Innovation, Nanyang Technological University (IMI-NTU). Any opinions, findings and conclusions or recommendations expressed in this material are those of the author(s) and do not reflect the views of National Research Foundation, Singapore. 
We would like to thank Bright Hill Evergreen Home and Goshen Consultants in Singapore for selecting the elderly and giving us access to the nursing home for this research.  
We would also like to thank our colleagues Yiep Soon and Ashwini Lawate for their support to setup Nadine for the experiment.

\bibliographystyle{IEEEtran}
\bibliography{bib}

\begin{thebibliography}{10}
\providecommand{\url}[1]{#1}
\csname url@samestyle\endcsname
\providecommand{\newblock}{\relax}
\providecommand{\bibinfo}[2]{#2}
\providecommand{\BIBentrySTDinterwordspacing}{\spaceskip=0pt\relax}
\providecommand{\BIBentryALTinterwordstretchfactor}{4}
\providecommand{\BIBentryALTinterwordspacing}{\spaceskip=\fontdimen2\font plus
\BIBentryALTinterwordstretchfactor\fontdimen3\font minus
  \fontdimen4\font\relax}
\providecommand{\BIBforeignlanguage}[2]{{%
\expandafter\ifx\csname l@#1\endcsname\relax
\typeout{** WARNING: IEEEtran.bst: No hyphenation pattern has been}%
\typeout{** loaded for the language `#1'. Using the pattern for}%
\typeout{** the default language instead.}%
\else
\language=\csname l@#1\endcsname
\fi
#2}}
\providecommand{\BIBdecl}{\relax}
\BIBdecl

\bibitem{correia2017social}
F.~Correia, P.~Alves-Oliveira, S.~Petisca, A.~Paiva \emph{et~al.}, ``Social and
  entertainment robots for older adults,'' 2017.

\bibitem{agrigoroaie2018physiological}
R.~Agrigoroaie and A.~Tapus, ``Physiological differences depending on task
  performed in a 5-day interaction scenario designed for the elderly: A pilot
  study,'' in \emph{International Conference on Social Robotics}.\hskip 1em
  plus 0.5em minus 0.4em\relax Springer, 2018, pp. 192--201.

\bibitem{agrigoroaie2018outcome}
------, ``The outcome of a week of intensive cognitive stimulation in an
  elderly care setup: a pilot test,'' in \emph{2018 27th IEEE International
  Symposium on Robot and Human Interactive Communication (RO-MAN)}.\hskip 1em
  plus 0.5em minus 0.4em\relax IEEE, 2018, pp. 814--819.

\bibitem{NadineSR_CGI_2019}
M.~Ramanathan, N.~Mishra, and N.~Magnenat~Thalmann, ``Nadine humanoid social
  robotics platform,'' in \emph{Computer Graphics International
  Conference}.\hskip 1em plus 0.5em minus 0.4em\relax Springer, 2019, pp.
  490--496.

\bibitem{law2019developing}
M.~Law, C.~Sutherland, H.~S. Ahn, B.~A. MacDonald, K.~Peri, D.~L. Johanson,
  D.-S. Vajsakovic, N.~Kerse, and E.~Broadbent, ``Developing assistive robots
  for people with mild cognitive impairment and mild dementia: a qualitative
  study with older adults and experts in aged care,'' \emph{BMJ open}, vol.~9,
  no.~9, p. e031937, 2019.

\bibitem{cohen2009engagement}
J.~Cohen-Mansfield, M.~Dakheel-Ali, and M.~S. Marx, ``Engagement in persons
  with dementia: the concept and its measurement,'' \emph{The American journal
  of geriatric psychiatry}, vol.~17, no.~4, pp. 299--307, 2009.

\bibitem{martinez2020socially}
E.~Martinez-Martin, F.~Escalona, and M.~Cazorla, ``Socially assistive robots
  for older adults and people with autism: An overview,'' \emph{Electronics},
  vol.~9, no.~2, p. 367, 2020.

\bibitem{thompson2017robot}
C.~Thompson, S.~Mohamed, W.-Y.~G. Louie, J.~C. He, J.~Li, and G.~Nejat, ``The
  robot tangy facilitating trivia games: A team-based user-study with long-term
  care residents,'' in \emph{2017 IEEE international symposium on robotics and
  intelligent sensors (IRIS)}.\hskip 1em plus 0.5em minus 0.4em\relax IEEE,
  2017, pp. 173--178.

\bibitem{li2016user}
J.~Li, W.-Y.~G. Louie, S.~Mohamed, F.~Despond, and G.~Nejat, ``A user-study
  with tangy the bingo facilitating robot and long-term care residents,'' in
  \emph{2016 IEEE international symposium on robotics and intelligent sensors
  (IRIS)}.\hskip 1em plus 0.5em minus 0.4em\relax IEEE, 2016, pp. 109--115.

\bibitem{louie2020social}
W.-Y.~G. Louie and G.~Nejat, ``A social robot learning to facilitate an
  assistive group-based activity from non-expert caregivers,''
  \emph{International Journal of Social Robotics}, pp. 1--18, 2020.

\bibitem{louie2015tangy}
W.-Y.~G. Louie, J.~Li, C.~Mohamed, F.~Despond, V.~Lee, and G.~Nejat, ``Tangy
  the robot bingo facilitator: a performance review,'' \emph{Journal of Medical
  Devices}, vol.~9, no.~2, 2015.

\bibitem{stevie2020bingo}
\BIBentryALTinterwordspacing
B.~C. Studies. (2020) Meet stevie the social robot that holds bingo lessons in
  a care home. [Online]. Available:
  \url{https://businesscasestudies.co.uk/meet-stevie-the-social-robot-that-holds-bingo-lessons-in-a-care-home/}
\BIBentrySTDinterwordspacing

\bibitem{khosla2013embodying}
R.~Khosla and M.-T. Chu, ``Embodying care in matilda: an affective
  communication robot for emotional wellbeing of older people in australian
  residential care facilities,'' \emph{ACM Transactions on Management
  Information Systems (TMIS)}, vol.~4, no.~4, pp. 1--33, 2013.

\bibitem{chang2013use}
W.~{Chang}, S.~{Šabanovic}, and L.~{Huber}, ``Use of seal-like robot paro in
  sensory group therapy for older adults with dementia,'' in \emph{2013 8th
  ACM/IEEE International Conference on Human-Robot Interaction (HRI)}, 2013,
  pp. 101--102.

\bibitem{vsabanovic2016socializing}
S.~{\v{S}}abanovi{\'c} and W.-L. Chang, ``Socializing robots: constructing
  robotic sociality in the design and use of the assistive robot paro,''
  \emph{AI \& society}, vol.~31, no.~4, pp. 537--551, 2016.

\bibitem{vsabanovic2013paro}
S.~{\v{S}}abanovi{\'c}, C.~C. Bennett, W.-L. Chang, and L.~Huber, ``Paro robot
  affects diverse interaction modalities in group sensory therapy for older
  adults with dementia,'' in \emph{2013 IEEE 13th international conference on
  rehabilitation robotics (ICORR)}.\hskip 1em plus 0.5em minus 0.4em\relax
  IEEE, 2013, pp. 1--6.

\bibitem{silbot2019blond}
\BIBentryALTinterwordspacing
L.~Blond, ``Studying robots outside the lab: Hri as ethnography,''
  \emph{Paladyn, Journal of Behavioral Robotics}, vol.~10, no.~1, pp. 117--127,
  2019. [Online]. Available: \url{https://doi.org/10.1515/pjbr-2019-0007}
\BIBentrySTDinterwordspacing

\bibitem{melkas2020impacts}
H.~Melkas, L.~Hennala, S.~Pekkarinen, and V.~Kyrki, ``Impacts of robot
  implementation on care personnel and clients in elderly-care institutions,''
  \emph{International Journal of Medical Informatics}, vol. 134, p. 104041,
  2020.

\bibitem{tangy}
J.~{Li}, W.~G. {Louie}, S.~{Mohamed}, F.~{Despond}, and G.~{Nejat}, ``A
  user-study with tangy the bingo facilitating robot and long-term care
  residents,'' in \emph{2016 IEEE International Symposium on Robotics and
  Intelligent Sensors (IRIS)}, 2016, pp. 109--115.

\bibitem{huisman2019two}
C.~Huisman and H.~Kort, ``Two-year use of care robot zora in dutch nursing
  homes: An evaluation study,'' in \emph{Healthcare}, vol.~7, no.~1.\hskip 1em
  plus 0.5em minus 0.4em\relax Multidisciplinary Digital Publishing Institute,
  2019, p.~31.

\bibitem{tuisku2019robots}
O.~Tuisku, S.~Pekkarinen, L.~Hennala, and H.~Melkas, ``Robots do not replace a
  nurse with a beating heart,'' \emph{Information Technology \& People}, 2019.

\bibitem{baka2019talking}
E.~Baka, A.~Vishwanath, N.~Mishra, G.~Vleioras, and N.~Magnenat~Thalmann,
  ``“am i talking to a human or a robot?”: A preliminary study of human’s
  perception in human-humanoid interaction and its effects in cognitive and
  emotional states,'' in \emph{Computer Graphics International
  Conference}.\hskip 1em plus 0.5em minus 0.4em\relax Springer, 2019, pp.
  240--252.

\bibitem{xiao2014human}
Y.~Xiao, Z.~Zhang, A.~Beck, J.~Yuan, and D.~Thalmann, ``Human--robot
  interaction by understanding upper body gestures,'' \emph{Presence:
  teleoperators and virtual environments}, vol.~23, no.~2, pp. 133--154, 2014.

\bibitem{beck2016motion}
A.~Beck, Z.~Zhijun, and N.~Magnenat~Thalmann, ``Motion control for social
  behaviors,'' in \emph{Context Aware Human-Robot and Human-Agent
  Interaction}.\hskip 1em plus 0.5em minus 0.4em\relax Springer, 2016, pp.
  237--256.

\bibitem{magnenat2014social}
N.~Magnenat~Thalmann and Z.~Zhang, ``Social robots and virtual humans as
  assistive tools for improving our quality of life,'' in \emph{2014 5th
  International Conference on Digital Home}.\hskip 1em plus 0.5em minus
  0.4em\relax IEEE, 2014, pp. 1--7.

\bibitem{mishra2019can}
N.~Mishra, M.~Ramanathan, R.~Satapathy, E.~Cambria, and N.~Magnenat~Thalmann,
  ``Can a humanoid robot be part of the organizational workforce? a user study
  leveraging sentiment analysis,'' in \emph{2019 28th IEEE International
  Conference on Robot and Human Interactive Communication (RO-MAN)}.\hskip 1em
  plus 0.5em minus 0.4em\relax IEEE, 2019, pp. 1--7.

\bibitem{neven2010but}
L.~Neven, ``‘but obviously not for me’: robots, laboratories and the
  defiant identity of elder test users,'' \emph{Sociology of health \&
  illness}, vol.~32, no.~2, pp. 335--347, 2010.

\bibitem{lee2016robot}
H.~R. Lee, H.~Tan, and S.~{\v{S}}abanovi{\'c}, ``That robot is not for me:
  Addressing stereotypes of aging in assistive robot design,'' in \emph{2016
  25th IEEE International Symposium on Robot and Human Interactive
  Communication (RO-MAN)}.\hskip 1em plus 0.5em minus 0.4em\relax IEEE, 2016,
  pp. 312--317.

\bibitem{chang2014exploring}
W.-L. Chang and S.~{\v{S}}abanovi{\'c}, ``Exploring taiwanese nursing homes as
  product ecologies for assistive robots,'' in \emph{2014 IEEE International
  Workshop on Advanced Robotics and its Social Impacts}.\hskip 1em plus 0.5em
  minus 0.4em\relax IEEE, 2014, pp. 32--37.

\bibitem{raft2020}
Z.~Teed and J.~Deng, ``Raft: Recurrent all-pairs field transforms for optical
  flow,'' in \emph{Proceedings of the IEEE Conference on Computer Vision and
  Pattern Recognition}, 2020.

\bibitem{ResNet}
K.~He, X.~Zhang, S.~Ren, and J.~Sun, ``Deep residual learning for image
  recognition,'' in \emph{Proceedings of European Conference on Computer
  Vision}, 2016, pp. 770--778.

\bibitem{celebA2018}
Z.~Liu, P.~Luo, X.~Wang, and X.~Tang, ``Large-scale celebfaces attributes
  (celeba) dataset,'' \emph{Retrieved August}, vol.~15, p. 2018, 2018.

\bibitem{TSN}
L.~Wang, Y.~Xiong, Z.~Wang, Y.~Qiao, D.~Lin, X.~Tang, and L.~Van~Gool,
  ``Temporal segment networks for action recognition in videos,'' \emph{IEEE
  transactions on pattern analysis and machine intelligence}, vol.~41, no.~11,
  pp. 2740--2755, 2018.

\end{thebibliography}
\end{document}